\documentclass[letterpaper, conference]{IEEEtran}

\IEEEoverridecommandlockouts                              

\usepackage{subcaption}
\usepackage{hyperref}       
\usepackage{booktabs}       
\usepackage{graphicx}
\usepackage[acronym,shortcuts]{glossaries}
\usepackage{comment}
\usepackage{enumerate}
\usepackage[capitalise]{cleveref}
\usepackage[table,xcdraw]{xcolor}
\usepackage{siunitx}
\usepackage{bm} 
\usepackage{amssymb} 
\usepackage{amsmath}
\usepackage[english]{babel}
\usepackage{multirow}
\usepackage{pifont}
\usepackage{eso-pic}

\newcommand\AtPageUpperMyleft[1]{\AtPageUpperLeft{%
\put(\LenToUnit{1cm},\LenToUnit{-2cm}){#1}%
}}%

\AddToShipoutPictureBG*{%
  \AtPageUpperMyleft{\parbox[b][2cm][c]{\paperwidth}{%
    \centering
    \fontsize{12}{14}\selectfont
    \color{gray!50}
    This paper has been accepted for publication at the\\
    International Conference on Intelligent Robots and Systems (IROS), Abu Dhabi 2024. \copyright{}IEEE
  }}%
}

\AddToShipoutPictureBG*{
  \AtPageLowerLeft{%
    \raisebox{25pt}{\makebox[\paperwidth]{\begin{minipage}{21cm}\centering
    \fontsize{10}{12}\selectfont
      \textcolor{gray!50}{ \copyright 2024 IEEE.  Personal use of this material is permitted.  Permission from IEEE must be obtained for all other uses, in any current or future media, including reprinting/republishing this material for advertising or promotional purposes, creating new collective works, for resale or redistribution to servers or lists, or reuse of any copyrighted component of this work in other works.
      }
    \end{minipage}}}%
  }
}

\definecolor{lightgray}{gray}{0.9}

\def\BibTeX{{\rm B\kern-.05em{\sc i\kern-.025em b}\kern-.08em
    T\kern-.1667em\lower.7ex\hbox{E}\kern-.125emX}}
\newcommand{\cmark}{\ding{51}}%
\newcommand{\xmark}{\ding{55}}

\title{\LARGE \bf \vspace{6mm}
CR3DT: Camera-RADAR Fusion for 3D Detection and Tracking
}

\author{Nicolas Baumann\IEEEauthorrefmark{1}, Michael Baumgartner\IEEEauthorrefmark{1}, Edoardo Ghignone\IEEEauthorrefmark{1}, Jonas Kühne\IEEEauthorrefmark{1},\\ Tobias Fischer\IEEEauthorrefmark{2}, Yung-Hsu Yang\IEEEauthorrefmark{2}, Marc Pollefeys\IEEEauthorrefmark{2}, and Michele Magno\IEEEauthorrefmark{1} \\
\thanks{\IEEEauthorrefmark{1}Nicolas Baumann, Michael Baumgartner, Edoardo Ghignone, Jonas Kühne, and Michele Magno are associated with the Center for Project-Based Learning, D-ITET, ETH Zurich.}%
\thanks{\IEEEauthorrefmark{2}Tobias Fischer, Yung-Hsu Yang, and Marc Pollefeys are associated with the Computer Vision and Geometry Group, D-INFK, ETH Zurich.}%
\thanks{Nicolas Baumann, Michael Baumgartner, Edoardo Ghignone, and Jonas Kühne contributed equally to this work. Corresponding author: Jonas Kühne.}
}

\begin{document}
\newacronym{lidar}{LiDAR}{Light Detection and Ranging}
\newacronym{radar}{RADAR}{Radio Detection and Ranging}
\newacronym{ml}{ML}{Machine Learning}
\newacronym{if}{IF}{Intermediate Frequency}
\newacronym{aoa}{AoA}{Angle of Arrival}
\newacronym{cfar}{CFAR}{Constant False Alarm Rate}
\newacronym{iot}{IoT}{Internet of Things}
\newacronym{bev}{BEV}{Bird's-Eye View}
\newacronym{sota}{SotA}{State-of-the-Art}
\newacronym{cr3dt}{CR3DT}{Camera-RADAR 3D Detection and Tracking}
\newacronym{cr3d}{CR3D}{Camera-RADAR 3D Detector}
\newacronym{cnn}{CNN}{Convolutional Neural Network}
\newacronym{lss}{LSS}{Lift Splat Shoot}
\newacronym{rgb}{RGB}{Red Green Blue}
\newacronym{map}{mAP}{mean Average Precision}
\newacronym{fpn}{FPN}{Feature Pyramid Network}
\newacronym{iou}{IOU}{Intersection over Union}
\newacronym{nds}{NDS}{nuScenes Detection Score}
\newacronym{mate}{mATE}{mean Average Translation Error}
\newacronym{fov}{FoV}{Field of View}
\newacronym{vfe}{VFE}{Voxel Feature Encoding}
\newacronym{amota}{AMOTA}{Average Multi-Object Tracking Accuracy}
\newacronym{amotp}{AMOTP}{Average Multi-Object Tracking Precision}
\newacronym{mot}{MOT}{Multi-Object Tracking}
\newacronym{ids}{IDS}{ID Switches}
\newacronym{roi}{ROI}{Region of Interest}
\newacronym{nms}{NMS}{Non-Maximum Suppression}
\newacronym{mave}{mAVE}{mean Average Velocity Error}
\newacronym{cbgs}{CBGS}{Class-Balanced Grouping and Sampling}
\newacronym{gpu}{GPU}{Graphics Processing Unit}
\newacronym{fps}{FPS}{Frames Per Second}
\newacronym{kf}{KF}{Kalman Filter}

\maketitle
\thispagestyle{empty}
\pagestyle{empty}

\begin{abstract}
To enable self-driving vehicles accurate detection and tracking of surrounding objects is essential. While \gls{lidar} sensors have set the benchmark for high-performance systems, the appeal of camera-only solutions lies in their cost-effectiveness. Notably, despite the prevalent use of \gls{radar} sensors in automotive systems, their potential in 3D detection and tracking has been largely disregarded due to data sparsity and measurement noise. As a recent development, the combination of \glspl{radar} and cameras is emerging as a promising solution.
This paper presents \gls{cr3dt}, a camera-\gls{radar} fusion model for 3D object detection, and \gls{mot}. Building upon the foundations of the \gls{sota} camera-only \emph{BEVDet} architecture, \gls{cr3dt} demonstrates substantial improvements in both detection and tracking capabilities, by incorporating the spatial and velocity information of the \gls{radar} sensor. Experimental results demonstrate an absolute improvement in detection performance of 5.3\% in \gls{map} and a 14.9\% increase in \gls{amota} on the nuScenes dataset when leveraging both modalities. \gls{cr3dt} bridges the gap between high-performance and cost-effective perception systems in autonomous driving, by capitalizing on the ubiquitous presence of \gls{radar} in automotive applications. The code is available at: \url{https://github.com/ETH-PBL/CR3DT}.
\end{abstract}

\setlength{\tabcolsep}{3pt}
\section{Introduction}
Perceiving and tracking the local surroundings is a pivotal task in the field of autonomous driving \cite{bevdet, caesar2020nuscenes, cc3dt}.
This has led to the development of a multitude of complex and high-performance 3D object detection and tracking architectures, primarily designed to perform on well-established datasets such as \emph{KITTI}, \emph{Waymo}, or \emph{nuScenes} \cite{kitti, waymo, caesar2020nuscenes}.
Recent 3D object detection methods mainly utilize two distinct sensor setups:

\begin{enumerate}[I]
\item \textbf{LiDAR-based methods for maximum performance:} Models following this paradigm lean heavily on the \gls{lidar} sensor modality \cite{pointpillars, liang2022bevfusion, centerpoint} to reach high accuracy scores, requiring high computational power and incurring high costs for the sensors and processing unit.
\item \textbf{Camera-only methods for cost-effectiveness:} Substituting the expensive but highly performant \gls{lidar} sensor with multiple cameras not only significantly reduces cost, but might allow for wider adoption of self-driving technology within the automotive industry \cite{li2022bevformer, cc3dt} while delivering competitive performance.
\end{enumerate}

While both strategies have their merits, a distinct performance gap exists between camera-only and \gls{lidar}-based models. Notably, in 3D detection tasks, \gls{sota} camera-only models achieve an \gls{map} of 62.4\,\% \cite{hop2023}, whilst \gls{sota} \gls{lidar}-only models reach 69.5\,\% \cite{lu2023link}. Similarly, \gls{sota} camera-only methods for 3D tracking achieve 65.3\% \gls{amota} \cite{wang2023streampetr}, whilst \gls{lidar}-based methods score 71.5\% \gls{amota} \cite{focalformer3d}.
It is worth noting that these impressive performance numbers are reached by leveraging high-resolution image inputs, incorporating temporal information, or utilizing offline detections, as indicated in \cref{tab:detection}. The latter implies that the model utilizes future data, rendering such detection systems impractical for real-time, i.e., online applications. These techniques may be used in various combinations, making fair comparisons about model performance difficult (see \cref{tab:detection}). This work targets the challenging task of object perception and tracking in autonomous driving and hence uses a computationally feasible, and fully online model that does not use additional temporal information (i.e., data corresponding to previous frames), to which we refer as the \emph{restricted model class}, further detailed in \cref{sec:expsetup}.

\begin{figure*}[t]
    \centering
    \includegraphics[trim={0.5cm 0 0 0.5cm},clip,width=\textwidth]{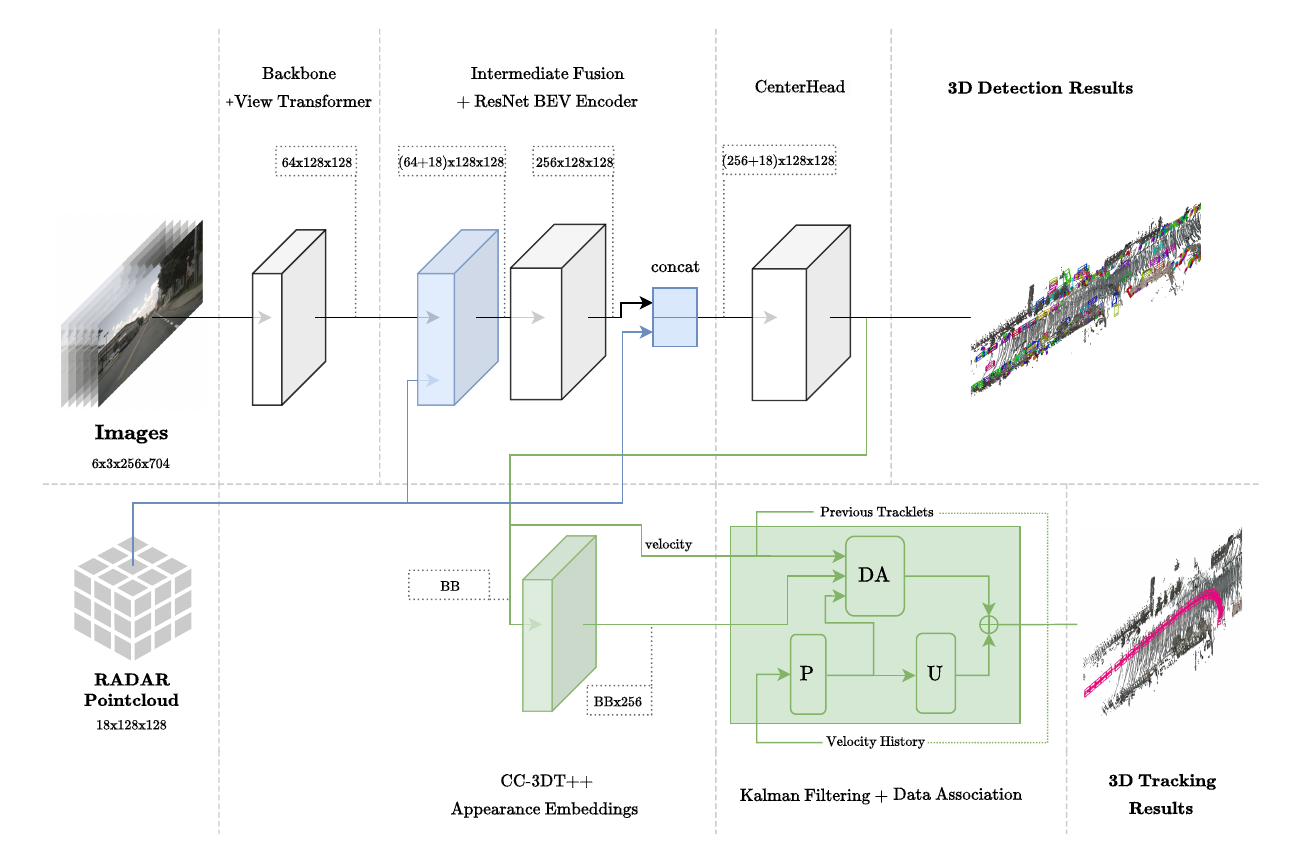}
     \caption{\textbf{Architectural Overview.} The model presented in this work extends \emph{BEVDet} \cite{bevdet}. Detection and tracking contributions are highlighted in light blue and green, respectively. Model inputs and outputs are highlighted in bold, with the input from the six different camera views being RGB images with resolution $704 \times 256$, and the radar input being encoded in a $128 \times 128$ \acrshort{bev} grid, in which each grid cell contains 18 features. The image stream is processed through a \emph{ResNet-50} backbone and then projected into \acrshort{bev} space utilizing a \gls{lss} view transformer, while the \gls{radar} stream is encoded in pillars with the feature encodings detailed in \cref{subsec:sensor_fusion}. The two intermediate outputs are then concatenated and processed through a \emph{ResNet} \acrshort{bev} Encoder as in \cite{bevdet}. After this step, the rasterized \gls{radar} pillars are concatenated once more through a residual connection. 
    The generated \acrshort{bev} features are then passed through a \emph{CenterPoint} detection head \cite{centerpoint}, which generates the detection result. The output bounding boxes are utilized to select \glspl{roi} and extract appearance embeddings \cite{cc3dt, qd3dt}. Finally, these embeddings are used in the \emph{Data Association (DA)} step, which generates the tracking results leveraging a refined velocity estimate together with the velocity output from the detection module. Such an estimate is obtained with a \gls{kf}, as in the \emph{KF3D} setting of \cite{cc3dt}, a two-stage state estimation that consists of a \emph{Prediction Step (P)} and an \emph{Updating Step (U)}. Green-dotted lines represent stored data saved across timesteps.}
    \label{fig:cr3dtarch}
\end{figure*}

The \gls{radar} sensor, prevalent in the automotive industry, has only recently emerged as a promising modality to bridge the existing performance gap. Previously, the sensor readings were deemed too noisy and sparse to improve 3D detection and tracking tasks \cite{caesar2020nuscenes, aw_radarcamera_survey}. The absence of \gls{radar}-based solutions in the \emph{nuScenes} tracking challenge \cite{CR_nuScenes_tracking}, a dataset that provides \gls{radar} data, highlights this fact. While the sensor readings offer a spatial point cloud akin to \glspl{lidar}, although sparser, they also deliver richer measurements, capturing data points that include velocity and \gls{radar}-reflectivity information. Only recently, \gls{radar} has been used to support tasks such as 3D segmentation \cite{simplebev} and 3D detection \cite{kim2023crn, xiong2023lxl}. Furthermore, different works have highlighted the \glspl{radar} increased robustness to adverse weather conditions \cite{aw_radarcamera_survey, aw_survey, hrfuser}, especially when compared to \glspl{lidar} and cameras. 

This paper aims to bridge the performance gap between \gls{lidar} and camera-only methods, by leveraging sensor fusion of the camera and \gls{radar} modality. Thus, we introduce \gls{cr3dt} a 3D detection and tracking solution synthesizing camera and \gls{radar} data, capitalizing on the added velocity data of the \gls{radar} readings. We opted to fuse these two types of sensor data within the \gls{bev} space, based on promising results demonstrated by using \gls{radar} in this domain \cite{simplebev, kim2023crn, xiong2023lxl}. Additionally, the \gls{bev} space has yielded high-performance outcomes in both camera-only techniques \cite{bevdet, li2022bevformer, li2023bevdepth, huang2022bevdet4d} and camera-\gls{lidar} fusion methods \cite{liang2022bevfusion, liu2022bevfusion}. Furthermore, the \gls{bev} space as an intermediate representation facilitates the incorporation of the 3D data from the \gls{radar}, a notable advantage over other works that relied on 2D projections \cite{hrfuser, aw_betz_radarcamera}. The main points of this work are summarized below:
\begin{itemize}
    \item \textbf{Sensors fusion architecture:} The proposed \gls{cr3dt} architecture integrates \gls{radar} data using intermediate fusion both before and after the \gls{bev} encoding head. It utilizes a quasi-dense appearance embedding head for tracking, trained similarly to \cite{cc3dt, qd3dt}. Additionally, the tracker explicitly uses the velocity estimates of the detector for the object association. 
    The model further detailed in \cref{fig:cr3dtarch}, represents one of the first tracking architectures that leverage the complementary camera and \gls{radar} modalities.
    \item \textbf{Detection performance evaluation:} \gls{cr3dt} achieves an \gls{map} of 35.1\,\% and a \gls{nds} of 45.6\,\% on the \emph{nuScenes} 3D detection validation set. This outperforms \gls{sota} single-frame, camera-only detection models within the \emph{restricted model class} as defined in \cref{sec:expsetup} by 5.3\,\% \gls{map} points. With the rich velocity information contained in the \gls{radar} data, the detector furthermore reduces the \gls{mave} by 45.3\,\% versus the previously mentioned \gls{sota} camera-only detector.
    \item \textbf{Tracking performance evaluation:} \gls{cr3dt} demonstrates a tracking performance of 38.1\% \gls{amota} on the \emph{nuScenes} tracking validation set. This corresponds to a 14.9\% \gls{amota} points improvement against \gls{sota} camera-only tracking models in the \emph{restricted model class} as defined in \cref{sec:expsetup}. The explicit use of velocity information and further improvements in the tracker significantly reduces the number of \gls{ids} by about 43\,\% versus the mentioned \gls{sota} model.
\end{itemize}

\section{Related Work}
Popular autonomous driving datasets such as \emph{KITTI} \cite{kitti}, \emph{Waymo} \cite{waymo}, and \emph{nuScenes} \cite{caesar2020nuscenes} allow model performance comparison of perception systems in the tasks of 3D object detection and tracking. Recent successful models typically leverage the \gls{bev} space to tackle these problems. In the following, we discuss previous \gls{lidar}-based models, camera-only models, and camera-\gls{radar} models addressing 3D object detection. Finally, we discuss the \gls{sota} models operating in 3D \gls{bev} space to perform object tracking.

\subsection{LiDAR-Based 3D Object Detection}
One of the first works that successfully used \gls{lidar} data in the 3D object detection task was \emph{VoxelNet} \cite{voxelnet}, which managed to encode sparse point cloud data into voxels thanks to the introduction of the novel \gls{vfe} layer.
Successively, \emph{SECOND} \cite{second} tried to remedy some of the performance problems that \emph{VoxelNet} suffered from -- mainly due to the incorporation of a 3D convolution module in its \gls{bev} encoding layer.
\emph{PointPillars} \cite{pointpillars} then was a seminal work that built upon \emph{VoxelNet} and \emph{SECOND}, removing the 3D convolution and encoding the point cloud features directly into pillars instead of voxels.
More recent 3D object detection approaches like \emph{CenterPoint} \cite{centerpoint} typically utilize either \emph{VoxelNet} or \emph{PointPillars} in their feature extraction backbone while improving on the object detection module. \emph{CenterPoint} itself introduced a highly effective two-stage detection architecture, which in the first stage extracts a rotation-agnostic heatmap of object positions, for which then in the second stage the bounding boxes are regressed.
It is noteworthy, that these methodologies and architectures are inherently sensor-agnostic and are thus also interesting for utilization with \gls{radar} point cloud data as well as \emph{lifted} image features in \gls{bev} space, as used in \gls{cr3dt}.

\subsection{Camera-Only 3D Object Detection}
The search for 3D detection architectures has extensively focused on cost-effective camera-only models. When using camera-based systems in \gls{bev} space, 2D object detection techniques are being leveraged as much as possible due to their proven effectiveness. They are extended into the third dimension using multiple cameras spaced around the car. 
Recent development in this field yielded a technique called \gls{lss} \cite{liftsplatshoot}, a pioneering method that leverages the camera intrinsics and extrinsic to project 2D image features into the \gls{bev} space of the car. 
While \gls{lss} utilizes the camera parameters as well as a per pixel attention-style \gls{cnn} operation in \gls{bev} space, there exist other methods such as parameter-free lifting \cite{simplebev} or methods based on deformable attention \cite{li2022bevformer}. In this work, however, we restrict ourselves to the simpler \gls{lss} technique that is also utilized in the \gls{sota} \emph{BEVDet} series.
 
Recent models leverage this image-view transformation to employ detection or segmentation heads on the aggregated \gls{bev} feature space \cite{bevdet, li2022bevformer, huang2022bevdet4d}. They differ mostly in the way the image features are extracted and \emph{lifted}, as well as subsequently encoded in the aggregated \gls{bev} space to be prepared for the detection or segmentation head. The \emph{CenterPoint} architecture \cite{centerpoint}, while initially developed for the \gls{lidar} modality, has been proven to work well with image-based features in \gls{bev} space, as shown by \emph{BEVDet} \cite{bevdet}, which denotes the current \gls{sota} in camera-only online 3D object detection within the \emph{restricted model class} as defined in \cref{sec:expsetup}. Therefore, our proposed solution builds upon the \emph{BEVDet} architecture, additionally integrating the \gls{radar} modality, and improving performance in terms of \gls{map}, \gls{nds}, and \gls{mave}.

\subsection{Camera-RADAR Fusion Models}
Recent camera-\gls{radar} fusion architectures have demonstrated the potential of sensor-fusion within the 3D \gls{bev} space. \emph{SimpleBEV} \cite{simplebev} represents an impactful work in 3D object segmentation and \gls{bev}-based sensor fusion, where the \gls{radar} point clouds are \emph{rasterized} and then concatenated with the \gls{bev} image-view features obtained from a \emph{ResNet-101} backbone. This concatenation fuses the spatial \gls{radar} data with the feature-rich camera data after the projection of the image features into 3D \gls{bev} space. The proposed technique demonstrated an approximate 8\%-point increase in absolute \gls{iou} score for the \emph{nuScenes} segmentation task. 
Following a similar idea, the \emph{BEVDet4D} paradigm introduced in \cite{huang2022bevdet4d} shows the advantages of concatenating different sensor streams in \gls{bev} space, with the key difference being, that instead of \gls{radar} data previous camera \gls{bev} features are fused with the current ones.
Lastly, recent interest in cross-attention from the sensor-fusion community can be seen in \cite{kim2023crn}, where deformable cross-attention can be found as a competitive method for \gls{bev} space multi-modal fusion in the tasks of \gls{bev} segmentation and 3D detection and tracking. 
Hence, recent work strongly suggests that fusing \gls{radar} with camera data yields a cost-effective performance increase across different perception tasks for autonomous driving.
Following this trend, this work proposes a 3D detection and tracking architecture that evaluates different fusion strategies in \gls{bev} space and achieves significant improvement in detection and tracking performance metrics, when compared to the \gls{sota} baselines belonging to the same \emph{restricted model class}.

\subsection{Multi Object Tracking}
In the context of 3D \gls{mot}, numerous solutions have traditionally relied on the \gls{lidar} sensor modality due to its ability to provide a comprehensive 360-degree \gls{fov} \cite{centerpoint, tracker0, tracker1, tracker2}. In contrast, 3D \gls{bev} tracking systems based on camera-only detection models have been relatively rare \cite{cam_tracker1, cam_tracker0}. This scarcity is, in part, because camera-based approaches often extend 2D \gls{mot} techniques, which rely on tracking by detection, and utilize motion or appearance cues for association between frames. These 2D techniques inherently lack the capacity to leverage the rich spatial information provided by a 360-degree \gls{fov} sensor such as \gls{lidar}.

More recent works in camera-only tracking have begun to diverge from 2D \gls{mot} techniques, focusing on associating temporal image-feature embeddings of the objects being tracked \cite{qd3dt, qdtrack}. These methods capitalize on quasi-dense similarity learning to enable robust and precise tracking. One notable development in this line of research is the \emph{CC-3DT} \cite{cc3dt} model, which innovatively extends tracking capabilities to handle joint associations between different cameras, thereby enabling tracking tasks to be performed through cross-camera correlation. We exploit the tracker of \emph{CC-3DT} and enhance it, demonstrating in \cref{tab:track_ablation} that our tuning improves the out-of-the-box performance of the \emph{CC-3DT} tracker across all considered metrics.

\section{Model Architecture}
\label{sec:model_architecture}
A broad overview of the model architecture is depicted in \cref{fig:cr3dtarch}. It is inspired by the \emph{BEVDet} architecture \cite{bevdet} but proposes a novel fusion operation to expand the sensor modalities from camera-only to camera-\gls{radar}. Subsequently, the proposed \gls{cr3dt} architecture incorporates the \emph{CC-3DT++} tracking head, which explicitly uses the improved velocity estimations of the \gls{radar}-augmented detector in its data association.
By projecting the features of the six camera views and five \glspl{radar} positioned around the ego-car into the \gls{bev} space, we achieve the full 360-degree \gls{fov} as in \gls{lidar}-based tracking methods. 

\subsection{Sensor Fusion in BEV Space}\label{subsec:sensor_fusion}
\begin{figure}[t]
    \centering
    \includegraphics[trim={-0.1cm 0.3cm 0.5cm 0.2cm},clip,width=\columnwidth]{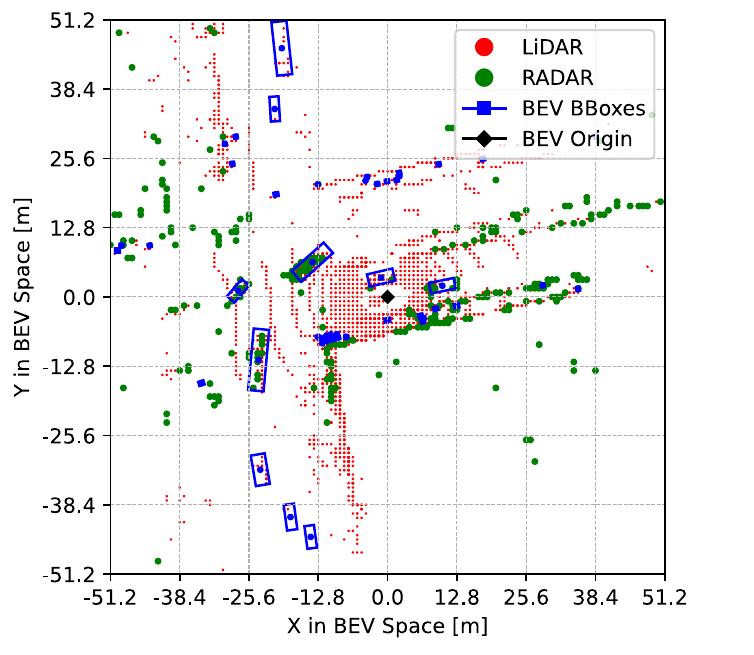}
     \caption{A visualization of the \gls{radar} point cloud aggregated into \gls{bev} space for the fusion operation. The \gls{lidar} point cloud is shown for visualization purposes only. The \gls{bev} space has a grid configuration of \SI{52.2}{\metre} with a resolution of \SI{0.8}{\metre}, resulting in a \texttt{(128x128)} dimensional feature space.}
    \label{fig:radpc}
\end{figure}

A visual representation of the \gls{radar} data within the same \gls{bev} space as the lifted image features can be seen in \cref{fig:radpc}. The image features however are not plotted for visibility reasons. Further, it is worth noting that the \gls{lidar} point cloud is solely used to visualize the surrounding \gls{bev} space more comprehensively for the reader and is not used for training of any kind. We utilize a \emph{PointPillars} \cite{pointpillars} inspired fusion method consisting of aggregation and concatenation without any reduction step, based on our findings in \cref{subsec:detector_ablation}. The \gls{bev} grid configuration is set to \SI{51.2}{\metre} range with a \SI{0.8}{\metre} resolution, which yields a feature grid of \texttt{(128x128)}. 

Similar to \emph{PointPillars}, the image features are lifted directly into 64-channel, single cell \emph{pillars} in the \gls{bev} grid, leading to an image feature size of \texttt{(64x128x128)}. The \gls{radar} data is also directly aggregated (employing averaging) into single pillars, utilizing all 18 \gls{radar} channels, including the point's x, y, and z locations. This implicitly adds the \emph{centroid} of the radar point cloud in each pillar to the list of features to be used. Notice that we did not augment the \gls{radar} data in any way, in contrast to \emph{PointPillars} augmentation efforts for \gls{lidar} point clouds \cite{pointpillars}. This is due to the fact that the \gls{radar} point cloud already includes more information compared to \gls{lidar}, as explored in \cite{simplebev}.
Subsequently, the \gls{radar} tensor of size \texttt{(18x128x128)} is concatenated to the image tensor of size \texttt{(64x128x128)} and both are fed into the \gls{bev} feature encoding layer as a \texttt{((64+18)x128x128)} tensor. Furthermore, the ablation study discussed in \cref{subsec:detector_ablation} showed, that it is beneficial to add a residual connection to the output of the feature encoding layer of dimension \texttt{(256x128x128)}, leading to a final input size to the \emph{CenterPoint} \cite{centerpoint} detection head of \texttt{((256+18)x128x128)}.

\subsection{Tracker Architecture} \label{subsec:tracker_arch}
The tracking architecture, which is integrated in \gls{cr3dt} and visible in \cref{fig:cr3dtarch}, is based on the \emph{CC-3DT} model\,\cite{cc3dt}. We introduce multiple technical adaptations in the data association step of the \gls{kf}, which significantly increase the tracking performance in terms of \gls{amota}, \gls{amotp}, and \gls{ids}.

The data association step is needed to associate the objects of two different frames and is based both on motion correlation and visual feature similarity. 
During training, 1D visual feature embedding vectors are obtained via quasi-dense multiple-positive contrastive learning as in \cite{qd3dt, qdtrack}.
Both the detections and feature embeddings are then used in the tracking stage of \emph{CC-3DT} \cite{cc3dt}, which is agnostic to the extraction process of the feature embeddings.
The data association step, referred to as \emph{DA} in \cref{fig:cr3dtarch}, was modified to leverage the improved positional detections and velocity estimates of \gls{cr3dt}. 
Specifically, the weighting terms and the formulation of the motion correlation matrix were redefined as detailed in the following paragraphs.

Following the naming convention in \emph{CC-3DT} \cite{cc3dt}, the detections $\mathcal{D}_t$ are associated with the active tracks of the \gls{kf} $\mathcal{T}_t$ at time $t$ with a greedy assignment given an affinity matrix $\mathbf{A}\left(\mathcal{T}_t, \mathcal{D}_t\right) \in \mathbb{R}^{\left\|\mathcal{T}_t\right\| \times \left\|\mathcal{D}_t\right\|}$.
The matrix is composed of the appearance embedding similarity matrix $\mathbf{A}_\texttt{deep}(\mathcal{T}_t,\, \mathcal{D}_t)$, the motion correlation matrix $\mathbf{A}_\texttt{motion}(\mathcal{T}_t,\, \mathcal{D}_t)$, and the location correlation matrix $\mathbf{A}_\texttt{loc}(\mathcal{T}_t,\, \mathcal{D}_t)$, weighted according to the following equation:
\begin{align}
\mathbf{A}(\mathcal{T}_t, \mathcal{D}_t) =& w_{\texttt{deep}} \mathbf{A}_\texttt{deep}(\mathcal{T}_t, \mathcal{D}_t) \nonumber\\ 
&+ w_{\texttt{motion}} \mathbf{A}_\texttt{motion}(\mathcal{T}_t, \mathcal{D}_t)\mathbf{A}_\texttt{loc}(\mathcal{T}_t, \mathcal{D}_t), 
\label{eq:affinityMatrix}
\end{align}
where $w_{\texttt{deep}}$ and $w_{\texttt{motion}}=1-w_{\texttt{deep}}$ are scalars. In this work, specific care was dedicated to the tuning of $w_{\texttt{motion}}$ to put more emphasis on the refined motion affinity terms.
The matrices $\mathbf{A}_{\texttt{deep}}$ and $\mathbf{A}_{\texttt{loc}}$ were left unchanged. The terms of the affinity matrix \( \mathbf{a}_\texttt{motion}(\tau_t,\, d_t) \) corresponding to a single track $\tau_t \in \mathcal{T}_t$ and detection $d_t \in \mathcal{D}_t$ are newly defined as:
\begin{equation}
    \mathbf{a}_\texttt{motion}(\tau_t, d_t) = \mathbf{a}_\texttt{vel} \mathbf{a}_\texttt{centroid} + (1 - \mathbf{a}_\texttt{vel}) \mathbf{a}_\texttt{pseudo}
\label{eq:motionCorrelation}
\end{equation}
where the centroid correlation \(\mathbf{a}_\texttt{centroid}\) and the state difference correlation \(\mathbf{a}_\texttt{pseudo}\) are defined as in \cite{cc3dt}. The new velocity correlation weight \( \mathbf{a}_\texttt{vel} \) is computed as:
\begin{equation}
    \mathbf{a}_\texttt{vel}(\tau_t, d_t) \!=\! \exp(-\frac{1}{r}|\mathbf{v}_{\tau_t} - \mathbf{v}_{d_t}|),
    \label{eq:correlationTerms}
\end{equation}
where $\mathbf{v}_{\tau_t}$ and $\mathbf{v}_{d_t}$ represent the subset of states related to the velocities (i.e., $\mathbf{v}_{s} = [v_x, v_y, v_z]^T, s \in \{\tau_t, d_t\}$) of a single track $\tau_t$ in the \gls{kf} and a detection $d_t$ respectively. These terms represent velocities, whereas in \cite{cc3dt} pseudo velocities based on the difference of the centroid position between two frames are used. Furthermore, in \cite{cc3dt} the pseudo velocities are compared by using the cosine similarity, whereas we use the velocity-based exponential function as shown in \cref{eq:correlationTerms}.

We refer to our extended implementation of \emph{CC-3DT} with adapted thresholds, weights, and the introduction of our novel velocity similarity weight as \emph{CC-3DT++}. The three changes are motivated by ablation studies, detailed in \cref{tab:track_ablation_tresh}, \cref{tab:track_ablation_weight}, and \cref{tab:track_ablation_velsim}, respectively.

\section{Experimental Setup}
\label{sec:expsetup}
This section details the setup and definitions of the detection and tracking baselines, as well as the training hardware that was utilized to train the described models. Within this work, both detection and tracking evaluations were performed on the well-known \emph{nuScenes} dataset, as it contains both the \gls{radar} and camera data, necessary for our method. We tested our model on the \emph{nuScenes} validation set, enabling comparison with related work. For computational reasons and to facilitate comparison, all models were trained without the usage of \gls{cbgs}.

\subsection{Detection Baseline}
Within this work, we built upon the well-established \emph{BEVDet} architecture \cite{bevdet} and refer to it as the detector baseline, or \gls{sota} camera-only model for single-frame inputs of the indicated image size. 
Due to computational resource limitations, all forthcoming 3D \gls{bev} detection performance results that involve \emph{BEVDet} based architectures are utilizing a \emph{ResNet-50} image encoding backbone and an image input size of \texttt{(3x256x704)}. Furthermore, only the current six input views per inference are used, i.e., neither past nor future temporal image information is utilized in any of our models, adhering to a fully online detection setting. We refer to this combination of settings as the \emph{restricted model class}.

\subsection{Training Hardware}
The training was performed on a single GeForce RTX 3090. To replicate the training of the \emph{BEVDet} network, which was performed on 8 \glspl{gpu} with a batch size of 8, training was performed with gradient accumulation over 8 steps. To alleviate the sparsity inherent to \gls{radar} data, five sweeps were used, i.e., the sweep associated with the current camera frames as well as the four previous \gls{radar} sweeps were accumulated. Note that the \gls{radar} sensor has a higher data rate than the camera, hence the radar sweeps that are being accumulated correspond to timestamps that are strictly after the timestamp of the previous image frame. No additional temporal information was used in our fusion models.

\section{Results} \label{sec:results}
For the detection results, the \gls{map}, \gls{nds}, and \gls{mave} scores are reported, while the tracking results use the \gls{amota}, \gls{amotp}, and \gls{ids} metrics. These scores are a subset of the official \emph{nuScenes} metrics, where \gls{nds} and \gls{amota} incorporate all the other metrics for detection and tracking respectively. For further explanation of the official \emph{nuScenes} metrics we refer to \cite{caesar2020nuscenes}.

\begin{table}[ht]
\resizebox{\columnwidth}{!}{%
\begin{tabular}{l|c|c|c|cccc}
\toprule
\textbf{Detection Models} & \textbf{Input} & \textbf{Resolution} & \textbf{Frames} & \textbf{\gls{map}} & \textbf{\gls{nds}} & \textbf{\gls{mave}} & \textbf{FPS} \\
& & & & \textbf{[\%]$\uparrow$} & \textbf{[\%]$\uparrow$} & \textbf{[$\frac{m}{s}$]$\downarrow$} & \textbf{[$\frac{1}{s}$]$\uparrow$} \\
\midrule
BEVDet (R50) \cite{bevdet} & C & 256$\times$704 & 1 & 29.8 & 37.9 & 0.86 & \textbf{30.2} \\ 
BEVFormerV2 $\dagger$ \cite{bevformerv2} & C & 256$\times$704 & 1 & 34.9 & 42.8 & 0.82 & - \\ 
CR3DT (\emph{ours}) & C+R & 256$\times$704 & 1 & \textbf{35.1} & \textbf{45.6} & \textbf{0.47} & 28.5 \\ 
\midrule
StreamPETR \cite{wang2023streampetr} & C & 512$\times$1408 & 8 & 50.4 & 59.2 & 0.26 & 6.4  \\ 
BEVDet (R101) \cite{bevdet} & C & 640$\times$1600 & 1 & 39.7 & 47.7 & 0.82 & 9.3 \\ 
BEVDet4D-Base \cite{huang2022bevdet4d} & C & 640$\times$1600 & 2 & 42.1 & 54.5 & \textbf{0.30} & 1.9  \\ 
HoP-BEVFormer \cite{hop2023} & C & 640$\times$1600 & 4 & 45.4 & 55.8 & 0.34 & - \\ 
CRN (R50) \cite{kim2023crn} & C+R & 256$\times$704 & 4 & 49.0 & 56.0 & 0.34 & \textbf{20.4} \\ 
CRN (R101) \cite{kim2023crn} & C+R & 512$\times$1408 & 4 & \textbf{52.5} & \textbf{59.2} & 0.35 & 7.2\\ 
\midrule
CenterPoint \cite{centerpoint} & L & - & 3 & \textbf{56.7} & \textbf{65.3} & - & -\\ 
\midrule
\multicolumn{8}{l}{$\dagger$: Offline 3D detections (i.e., using future information).} \\
\end{tabular}%
}
\caption{Detection results on the \emph{nuScenes} validation set. The top section lists detection models that conform to the resolution and temporal frame settings adopted in this work. The middle section includes models without such restrictions, granting them a significant advantage in detection performance. The last row contains a \gls{lidar}-based model for reference. \emph{C, R, L} denote the sensor modalities, camera, \gls{radar} and \gls{lidar}, respectively. \gls{fps} are reported from literature using an RTX 3090 \gls{gpu}.}
\label{tab:detection}
\end{table}

\subsection{Detection Results}\label{subsec:detect_res}
\Cref{tab:detection} shows the detection performance of \gls{cr3dt} compared to the baseline camera-only \emph{BEVDet} (R50) architecture. It is evident that the inclusion of the \gls{radar} sensor modality significantly increases the detection performance. Within the small resolution and temporal frame constraints, \gls{cr3dt} manages to achieve a 5.3\% \gls{map} percentage point improvement and a 7.7\% \gls{nds} percentage point improvement against the \gls{sota} camera-only \emph{BEVDet}. Even when compared to more complex offline architectures such as \emph{BEVFormerV2}, which uses a transformer-based attention module for perspective view supervision with bi-directional temporal encoders \cite{bevformerv2}, \gls{cr3dt} surpasses its performance in an online setting.

Furthermore, \cref{tab:detection} highlights the benefits of using higher resolutions and incorporating temporal frame information. Models operating under less constrained settings, i.e., higher image resolution or temporal frames, consistently outperform their restricted counterparts. This suggests the \gls{cr3dt} architecture could achieve better results if not limited by its current constraints, primarily computational ones. For instance, adapting the \gls{cr3dt} approach to the high-resolution \emph{BEVDet (R101)} might lead to a performance increase. Including temporal information could offer further improvements.

\cref{tab:detection} also shows the performance difference between \gls{lidar} and camera-only models. While \emph{CenterPoint} outperforms all camera-based models, integrating camera and \gls{radar} data narrows down this performance gap. The unrestricted \emph{CRN (R101)} comes within 4.2\% \gls{map} points of the \emph{CenterPoint} \gls{lidar} baseline, whereas the constrained \gls{cr3dt} is still 21.3\% \gls{map} points behind.

\begin{table}[ht]
    \begin{subtable}{.95\columnwidth}
        \centering
        \begin{tabular}{l|c|c|c}
            \toprule
            \textbf{Bins [\#]}  & \textbf{\gls{map} [\%]$\uparrow$} & \textbf{\gls{nds} [\%]$\uparrow$} & \textbf{\gls{mave} [$\frac{m}{s}$]$\downarrow$} \\
            \midrule
            1 & \textbf{34.66} & \textbf{45.14} & \textbf{0.45}  \\
            10 & 32.06  & 42.78 & 0.52  \\
            \bottomrule
        \end{tabular}
        \caption{Fusion Ablation 1: Effect of a $z$-dimension discretization and subsequent \emph{bev\_compressor} module within the sensor fusion step.}
        \label{tab:ablation_zdim}
    \end{subtable}
    \hfill
    \vspace{10pt}
    \begin{subtable}{.95\columnwidth}
        \centering
        \begin{tabular}{l|c|c|c}
            \toprule
            \textbf{Residual Connection} & \textbf{\gls{map} [\%]$\uparrow$} & \textbf{\gls{nds} [\%]$\uparrow$} & \textbf{\gls{mave} [$\frac{m}{s}$]$\downarrow$} \\
            \midrule
            Yes & \textbf{35.15} & \textbf{45.61} & 0.47  \\
            No & 34.66 & 45.14 & \textbf{0.45}\\
            \bottomrule
        \end{tabular}
        \caption{Fusion Ablation 2: Effect of a residual connection for the \gls{radar} data to the output of the intermediate fusion step as shown in \cref{fig:cr3dtarch}, without $z$-discretization.}
        \label{tab:ablation_latefus}
    \end{subtable}
    \caption{Detection architecture ablation experiments.}
    \label{tab:ablation}
\end{table}
\subsection{Camera-Radar Fusion Ablation}
\label{subsec:detector_ablation}
To analyze the camera-\gls{radar} fusion method in more detail, we compare different fusion architectures in \cref{tab:ablation}. Namely, we evaluate the performance of two different intermediate fusion approaches in \cref{tab:ablation_zdim} and the benefit of our residual connection in \cref{tab:ablation_latefus}.
Our intermediate fusion approaches were inspired by \emph{VoxelNet} \cite{voxelnet} and \emph{PointPillars} \cite{pointpillars}, respectively. The first approach includes a voxelization of the lifted RGB features and the pure \gls{radar} sensor data into cubes of dimension \SI{0.8}x\,\SI{0.8}x\,\SI{0.8}{\metre}, leading to an alternate feature size of \texttt{((64+18)x10x128x128)} and consequently to the necessity of a \emph{bev\_compressor} module in the form of a 3D convolution, similar to the one employed in \cite{voxelnet}. The latter approach is the one described in \cref{sec:model_architecture}, which forgoes the voxelization in the z dimension and the subsequent 3D convolution, and directly aggregates the lifted RGB features and pure \gls{radar} sensor data into pillars, leading to the known feature size of \texttt{((64+18)x128x128)}. As can be seen in \cref{tab:ablation_zdim}, omitting the discretization in the z dimension and the additional \emph{bev\_compressor} in the model architecture leads to an increase in the models \gls{map} and \gls{nds} as well as a decrease in its \gls{mave}.
Going further with the better model, we add a residual connection to the output of our initial camera-\gls{radar} fusion as seen in the system architecture overview in \cref{fig:cr3dtarch}. Such a residual connection leads to a further increase in the \gls{map} and \gls{nds} of our model, although at the cost of an increase in \gls{mave}, as can be seen in \cref{tab:ablation_latefus}. It is worth noting though, that the \gls{mave} still outclasses the camera-only \emph{BEVDet} base model, as seen in \cref{tab:detection}.

\begin{table*}[t]
    \centering
    {%
    \begin{tabular}{l|c|c|c|ccc|ccc}
    \toprule
    \textbf{Experiment Name} & \multicolumn{3}{|c|}{\textbf{Ablation Parameters}} &  \multicolumn{3}{|c|}{\textbf{BEVDet}: Tracking Performance } & \multicolumn{3}{|c}{\textbf{\gls{cr3dt}}: Tracking Performance } \\
    \midrule
     &\textbf{threshold} & \textbf{weight} & \textbf{vel. sim.} & \textbf{\gls{amota} [\%]$\uparrow$} & \textbf{\gls{amotp} [m]$\downarrow$} & \textbf{\gls{ids} [\#] $\downarrow$} & \textbf{\gls{amota} [\%]$\uparrow$} & \textbf{\gls{amotp} [m]$\downarrow$} & \textbf{\gls{ids} [\#] $\downarrow$}\\
    \midrule
    
     CC-3DT & \xmark & \xmark & \xmark & 23.2 (+0.0) & 1.48 (+0.00) &  2491 (+0) &  31.2 (+0.0) & 1.34 (+0.00) & 2809 (+0) \\ 
    \midrule[0.5\lightrulewidth]
    
     CC-3DT + Abl. 1&\cmark & \xmark & \xmark & 27.7 (+4.5) & 1.50 (+0.02) & 1140 (-1351) &  34.2 (+3.0) & 1.39 (+0.05) & \textbf{1291 (-1518)} \\ 
    \midrule[0.5\lightrulewidth]
    
     CC-3DT + Abl. 2&\xmark & \cmark & \xmark & 24.5 (+1.3) & \textbf{1.45 (-0.03)} & 2861 (+370) &  33.1 (+1.9) & \textbf{1.31 (-0.03)} & 3649 (+840) \\ 
    \midrule[0.5\lightrulewidth]
    
     CC-3DT + Abl. 1 \& 2&\cmark & \cmark & \xmark & 30.3 (+7.1) & 1.49 (+0.01) & 1122 (-1369) &  37.6 (+6.4) & 1.37 (+0.03) & 1537 (-1272) \\ 
    \midrule[0.5\lightrulewidth]
    
     CC-3DT++ (\emph{ours})&\cmark & \cmark & \cmark & \textbf{30.5 (+7.3)} & 1.49 (+0.01) & \textbf{1121 (-1370)} &  \textbf{38.1 (+6.9)} & 1.37 (+0.03) & 1432 (-1377) \\ 
    \bottomrule
    \end{tabular}%
    }
    \caption{
    Tracking results on the \emph{nuScenes} validation set for different tracker configurations based on both the baseline \emph{BEVDet} and our \gls{cr3dt} detector. The tracking architectures correspond to the baseline \emph{CC-3DT} tracker as well as the best intermediate models found in the ablation studies shown in \cref{tab:track_ablation} and our final \emph{CC-3DT++} tracker.
    We observe similar performance gains with both detection backbones for our general tracker improvements over the respective baseline. Although notably, \gls{cr3dt} benefits more from our novel velocity similarity term explained in \cref{subsec:tracker_arch}.}
    \label{tab:tracking}
    \end{table*}

\subsection{Tracking Results}
\Cref{tab:tracking} presents the tracking results of our improved \emph{CC-3DT++} tracking model discussed in \cref{subsec:tracker_arch} on the \emph{nuScenes} validation set. We report the performance of the tracker on top of both the baseline and \gls{sota} camera-only \emph{BEVDet} detector, as well as our \gls{cr3dt} detection model.
Alongside the results of the final \emph{CC-3DT++} tracking architecture, we also report the individual tracking performances utilizing the baseline \emph{CC-3DT} model \cite{cc3dt} as well as different intermediate tracking architectures explored in our tracking architecture ablation study in \cref{subsec:tracking_ablation}.

The proposed \gls{cr3dt} detector combined with the \emph{CC-3DT++} tracker architecture shows significant improvements in \gls{amota} and \gls{amotp} compared to the baseline camera-only \emph{BEVDet} detector and \emph{CC-3DT} tracker. Concretely, the individual improvements in the detector and tracker lead to a joint performance gain over the baseline of 14.9\,\% points in \gls{amota} and a reduction of \SI{0.11}{\metre} in \gls{amotp}. Furthermore, we see an \gls{ids} decrease of about 43\,\% compared to the baseline.

\subsection{CC-3DT++ Tracking Architecture Ablation}
\label{subsec:tracking_ablation}

To better understand the impact of our different changes to the original \emph{CC-3DT} \cite{cc3dt} tracking architecture and the addition of our velocity-based affinity term, we conduct extensive ablation studies depicted in \cref{tab:track_ablation}. Furthermore, we report the performance of the different tracking architecture configurations on both our baseline \emph{BEVDet} detector as well as our own \gls{cr3dt} detector in \cref{tab:tracking} to investigate the effects of our changes more generally.

\begin{table}[ht]
    \begin{subtable}{0.95\columnwidth}
        \centering
        \begin{tabular}{l|c|c}
            \toprule
            \textbf{Matching Score Threshold}  & \textbf{\gls{amota} [\%] $\uparrow$} & \textbf{\gls{ids} [\#] $\downarrow$} \\
            \midrule
            0.50 (default) & 32.3 & 2542 \\
            0.30 & \textbf{34.2}  & \textbf{1291}  \\
            0.18 & 29.5  & 1424  \\
            \bottomrule
        \end{tabular}
        \caption{Abl. 1: Effect of different matching score thresholds in the data association step of the tracker.}
        \label{tab:track_ablation_tresh}
    \end{subtable}%
    \hfill
    \vspace{10pt}
    \begin{subtable}{0.95\columnwidth}
        \centering
        \begin{tabular}{ll|c|c}
            \toprule
            \boldsymbol{$w_{\texttt{deep}}$} & \boldsymbol{$w_{\texttt{motion}}$}  & \textbf{\gls{amota} [\%] $\uparrow$} & \textbf{\gls{ids} [\#] $\downarrow$} \\
            \midrule
            0.5 & 0.5 (default) & 32.3 & \textbf{2542}  \\
            0.75 & 0.25 & 28.7  & 3134  \\
            0.25 & 0.75 & \textbf{33.1}  & 3649  \\
            \bottomrule
        \end{tabular}
        \caption{Abl. 2: Effect of various affinity matrix weights in \cref{eq:affinityMatrix}.}
        \label{tab:track_ablation_weight}
    \end{subtable}
    \hfill
    \vspace{10pt}
    \begin{subtable}{0.95\columnwidth}
        \centering
        \begin{tabular}{l|c|c}
            \toprule
            $\textbf{a}_\texttt{motion}$ \textbf{Tradeoff Parameter} & \textbf{\gls{amota} [\%] $\uparrow$} & \textbf{\gls{ids} [\#] $\downarrow$} \\
            \midrule
            cosine similarity (default) & 37.6 & 1537  \\
            velocity similarity (\emph{ours}) & \textbf{38.1} & \textbf{1432}  \\
            \bottomrule
        \end{tabular}
        \caption{Abl. 3: Effect of our newly introduced velocity similarity term as the trade-off term in \cref{eq:motionCorrelation} compared to the original \emph{CC-3DT} cosine similarity term. This is applied to the model with the best-shown threshold and affinity weights from the two previous ablations.}
        \label{tab:track_ablation_velsim}
    \end{subtable}
    \caption{Tracking architecture ablation experiments conducted on the \gls{cr3dt} detection backbone. All ablations include a bugfix in the original tracker architecture leading to a small performance gain in the default configuration.}
    \label{tab:track_ablation}
\end{table}

We ran three main experimental studies, the first was an investigation of the matching score threshold utilized in the greedy matching algorithm of \emph{CC-3DT}, which is tightly coupled to the affinity scores calculated in \cref{eq:affinityMatrix}. 
We observe significant performance gains both in an increase of \gls{amota} and a stark decrease in \gls{ids} for a slightly lower threshold, with a stark performance decrease for too low of a threshold, as indicated in \cref{tab:track_ablation_tresh}. 
Secondly, we explored different weightings of the embeddings correlation and motion correlation terms in \cref{eq:affinityMatrix}. 
As shown in \cref{tab:track_ablation_weight}, we find that a larger weight on the motion correlation term leads to an \gls{amota} gain, although at the cost of an \gls{ids} increase.
Lastly, we combined the best threshold and weighting configuration and examined the motion correlation term in \cref{eq:motionCorrelation} itself. Trying to leverage the much-improved velocity estimations of our detector, we exchanged the original trade-off parameter (the cosine similarity between the predicted motion direction and the observed motion direction in the xy-plane) with a new velocity similarity term, which compares the predicted with the observed velocity directly as explained in \cref{eq:correlationTerms}. As hypothesized, we observe a further performance increase in both \gls{amota} and \gls{ids} in \cref{tab:track_ablation_velsim}.

It additionally has to be noted, that we fixed a minor bug in the original implementation of \emph{CC-3DT}, leading to the performance increase observed in the default configurations shown in the ablations in comparison to \cref{tab:tracking}
(32.3\% \gls{amota} vs. 31.2\% \gls{amota})
.

Examining \cref{tab:tracking} now, we can draw more general conclusions regarding our refined tracker. First, we see a similar trend with both detection backbones concerning our general tuning improvements, underlining the merit of a proper analysis of matching scores within the tracker.
Secondly, and more interestingly, we see a performance gain on both backbones when introducing our new velocity correlation term into the matching score, albeit a smaller gain on the \emph{BEVDet} detector. 
This detail is particularly highlighted by the relative improvement in \gls{ids}: while the \emph{BEVDet}-based tracker shows almost no change (-0.1\%), the camera-\gls{radar} results show a significant decrease in \gls{ids} (-6.8\%) when the velocity similarity term is used.
This shows promise for such a correlation term in trackers in general and also highlights the positive effect of our improved velocity detections, due to the inclusion of \gls{radar} data in \gls{cr3dt}, for downstream tasks.

\subsection{Computational Results}
The computational results were obtained by measuring the inference time required for both the detection and tracking separately, on an RTX 3090 \gls{gpu}. The combined latency of the camera-only baseline \emph{BEVDet}+\emph{CC-3DT} (\SI{88.1}{\milli\second}, 11.35 \gls{fps}) is slightly faster than the proposed \gls{cr3dt} (\SI{90.04}{\milli\second}, 11.11 \gls{fps}). This is due to the additional computation of the \gls{radar} fusion. Hence, the inclusion of the \gls{radar} modality in \gls{cr3dt} accounts for a 2.2\% relative increase in latency compared to the camera-only baseline. However, note that the tracking inference is not optimized for latency and accounts for \SI{55}{\milli\second} in either the camera-only or \gls{cr3dt} pipeline. Hence the detectors, i.e., \emph{BEVDet} and \gls{cr3dt} alone, yield a latency of \SI{33.1}{\milli\second} (30.21 \gls{fps}) and \SI{35.04}{\milli\second} (28.54 \gls{fps}), respectively, as shown in \cref{tab:detection}.

\section{Conclusion}
This work presents \gls{cr3dt}, an efficient camera-\gls{radar} fusion model tailored to 3D object detection and \gls{mot}. By integrating \gls{radar} into the \gls{sota} camera-only \emph{BEVDet} architecture and introducing the \emph{CC-3DT++} tracking architecture, \gls{cr3dt} demonstrates a substantial increase in both detection and tracking accuracies — 5.35\% \gls{map} and 14.9\% \gls{amota} points, respectively.
This approach introduces a promising direction in the field of perception for autonomous driving, which leads to a cost-effective, performant, and \gls{lidar}-free system. It bridges the performance gap between the high-performance \gls{lidar}-based systems and the more cost-effective camera-only solutions.
While this study did not specifically investigate the \glspl{radar} inherent resilience to challenging weather conditions, the growing emphasis on \gls{radar} in recent research underscores its potential value in ensuring robust perception under adverse environmental scenarios. Therefore, future work could investigate potential robustness benefits of \gls{cr3dt}.





\section*{Acknowledgement}
We extend our gratitude to Dr. Christian Vogt of the Center for Project-Based Learning at ETH Zurich for his fruitful discussions and proofreading.


\bibliographystyle{IEEEtran}
\bibliography{main}

\end{document}